\documentclass{article}

\usepackage{float}
\usepackage{graphicx}
\usepackage[lofdepth,lotdepth]{subfig}
\usepackage{wrapfig,lipsum}
\usepackage{caption}

\usepackage{amsmath}
\usepackage{amssymb}
\usepackage{filecontents}

\usepackage[square,sort,comma,numbers]{natbib}

\usepackage{graphicx}

\usepackage[final]{nips_2017}

\usepackage[utf8]{inputenc} 
\usepackage[T1]{fontenc}    
\usepackage{hyperref}       
\usepackage{url}            
\usepackage{booktabs}       
\usepackage{amsfonts}       
\usepackage{nicefrac}       
\usepackage{microtype}      

\usepackage{array}
\usepackage{url}

\usepackage{lscape}

\title{Towards Symbolic Reinforcement Learning \\
with Common Sense}

\author{
  Artur d'Avila Garcez\textsuperscript{1}\\
  \And
  Aimore Resende Riquetti Dutra\textsuperscript{2}\\
  \And
  Eduardo Alonso\textsuperscript{3}\\
  \and\\
Department of Computer Science\\
City, University of London\\
London EC1V 0HB, U.K.\\
\texttt{a.garcez@city.ac.uk}\textsuperscript{1}, \texttt{aimorerrd@hotmail.com}\textsuperscript{2}, \texttt{e.alonso@city.ac.uk}\textsuperscript{3}
}

\begin{document}

\maketitle

\begin{abstract}
Deep Reinforcement Learning (deep RL) has made several breakthroughs in recent years in applications ranging from complex control tasks in unmanned vehicles to game playing. Despite their success, deep RL still lacks several important capacities of human intelligence, such as transfer learning, abstraction and interpretability.
Deep Symbolic Reinforcement Learning (DSRL) seeks to incorporate such capacities to deep Q-networks (DQN) by learning a relevant symbolic representation prior to using Q-learning. 
In this paper, we propose a novel extension of DSRL, which we call Symbolic Reinforcement Learning with Common Sense (SRL+CS), offering a better balance between generalization and specialization, inspired by principles of common sense when assigning rewards and aggregating Q-values.  
Experiments reported in this paper show that SRL+CS learns consistently faster than Q-learning and DSRL, achieving also a higher accuracy. In the hardest case, where agents were trained in a deterministic environment and tested in a random environment, SRL+CS achieves nearly 100\% average accuracy compared to DSRL's 70\% and DQN's 50\% accuracy.
To the best of our knowledge, this is the first case of near perfect zero-shot transfer learning using Reinforcement Learning.
\end{abstract}





\section{Introduction}
The combination of classical Reinforcement Learning (RL) with deep neural networks has achieved human-level competence at solving some difficult problems, especially with the use of Deep Q-Networks (DQN) at solving games\cite{Mnih2013,mnih2015human,silver2017mastering}. There is no doubt that Deep Reinforcement Learning (DRL) has offered new perspectives to the areas of automation and Artificial Intelligence (AI), but despite their success, DRL seems unable to tackle a number of problems that are considered relatively simple for humans to solve.  DRL requires large training data sets, thus learning slowly. A trained deep network performing very well in one task will often perform poorly in another, analogous (sometimes very similar) task. Finally, DRL lacks interpretability, being criticized frequently for being a ``black-box'' model.

Some authors have tried to address the above shortcomings by adding prior knowledge to neural networks in general, and more recently to RL \cite{garcez_book2, kaelbling1996reinforcement, lecun2015deep, son}. 
In \cite{garnelo2016towards}, it is argued that combining aspects of symbolic AI with neural networks and Reinforcement Learning could solve at once all of the above shortcomings. Such ideas are then instantiated in a computational system called Deep Symbolic Reinforcement Learning (DSRL). DSRL uses convolutional neural networks which, applied to navigation tasks typically consisting of negotiating obstacles (objects) and finding optimal paths or collecting other objects, can learn relevant object types from images, thus building a more abstract symbolic representation of the state-space, which is followed by a Q-learning algorithm for learning a policy on this state-space. Despite the fact that DSRL does not learn as effectively as DQN in a deterministic environment, DSRL was shown to outperform DQN when the policy learned in a deterministic environment is transferred to a random environment, thus indicating the value of using a more abstract, symbolic representation.

In this paper, we seek to reproduce the DSRL system with some simplifications. We have implemented this simplified, but also more generic system, which we refer to as Symbolic Reinforcement Learning (SRL). For the sake of analysis, it is desirable to separate within SRL, learning and decision-making from the object recognition part of the system, this latter part carried out in DSRL by convolutional neural networks. Such a separation allows us to focus the analysis of performance on the learning and decision making by providing the symbols and positions of the objects directly to the SRL system. Aspects of type transition and symbol interaction dealt with by DSRL are not analyzed in this paper.

We also propose an extension of SRL called Symbolic Reinforcement Learning with Common Sense (SRL+CS). SRL+CS makes one modification to the learning of SRL and one modification to the decision-making. This is then shown empirically to be sufficient for producing a considerable improvement in performance by comparison with various other algorithms. Differently from SRL, in SRL+CS only the states containing objects with which an agent interacts have their Q-values updated when rewards are non-zero (let us call this \emph{principle one}), and the relative position of an object with respect to an agent is considered in the decision of which action to take (call this \emph{principle two}).

More specifically, \emph{principle one} is implemented through Equation ( \ref{eq:SRL+CS_learn}) by making specific updates to the Q-values according to the specific interactions that an agent has with the (objects in the) environment; \emph{principle two}, which seeks to give more importance to the objects that are nearer to the agent, is implemented through Equation (\ref{eq:SRL+CS_act}) by simply dividing the Q-values by a factor proportional to the relative position of the agent w.r.t. the objects for the purpose of choosing an action. The implementation of these two principles within a standard RL set-up such as SRL+CS is only made possible by the adoption of a more abstract symbolic representation of the state-space, as put forward by the DSRL approach. 

For the sake of comparison between the various RL systems already mentioned, we chose to reproduce a benchmark game used by DSRL in \cite{garnelo2016towards}. In this game, SRL+CS achieves 100\% accuracy when tested in a random environment, compared to DQN's 50\% and DSRL's 70\% accuracy. In general, SRL+CS learns faster and more effectively than SRL by simply incorporating the above principles one and two to Q-learning.

The remainder of the paper is organized as follows: In Section 2, the Q-Learning, DQN and DSRL algorithms are reviewed briefly.
In Section 3, the Symbolic Reinforcement Learning (SRL) algorithm is explained in three parts: state-space representation, learning, and decision-making. In Section 4, the Symbolic Reinforcement Learning with Common Sense algorithm is introduced through simple modifications to the SRL algorithm's learning and decision making. The setup of the experimental environment is introduced in Section 5 using several variants of the benchmark game where an agent is expected to collect an object type while avoiding another. Experimental results are also  presented in Section 5. We evaluate how Q-learning, DQN, DSRL, SRL and SRL+CS generalize and scale with changes to the size of the environment, as well as how the systems perform at zero-shot transfer learning when trained and tested in deterministic or random setups. SRL+CS outperforms all the other systems in all but one of these experiments. We then conclude and discuss directions for future work.

\section{Background}

In this section, we briefly recall the algorithms of Q-learning, Deep Q-Networks (DQN) and Deep Symbolic Reinforcement Learning (DSRL).

\subsection{Q-Learning}

Q-learning is a model-free RL algorithm (although model-based variations have been proposed). In Q-learning, an agent's goal is to maximize by trial and error its total future reward thus learning which action is optimal in a given state \cite{watkins}. By learning a function $Q(s,a)$ mapping states to actions, Q-learning can under certain conditions find an optimal action-selection policy for any given finite Markov Decision Process (MDP). $Q(s,a)$ denotes the quality of a state-action pair, which ultimately gives the expected utility of an action $a$ in a given state $s$. The Q-learning algorithm can be summarized as follows (below, an $\epsilon$-greedy policy is the most commonly used policy, which with probability $\epsilon$, selects an action randomly, and with probability $1-\epsilon$, selects an action that gives the maximum reward in the current state):

\begin{tabular}{ll}
Starting at state: & $s_0$,\\
For & $t = 1,2,3,\dots$\\
& Choose an action $a_t$ using e.g. an $\epsilon$-greedy policy w.r.t. $Q(s,a);$\\
& Execute action $a_t$;\\
& Update Q, as follows:\\
& $Q(s_t,a_t) = Q(s_t,a_t) + \alpha_t\left[R(s_t,a_t) + \gamma\max_{a}Q(s_{t+1},a) - Q(s_t,a_t) \right]$,\\ 
& where $\alpha_t$ is the learning rate at time $t$,\\
& $R$ is the reward observed for the current state and choice of action,\\ 
& $\gamma$ is a temporal discount factor, and\\
& $\max_{a}Q(s_{t+1},a)$ is an estimate of the optimal future value.\\ 
\end{tabular}


At each step, the Q-Learning algorithm updates the Q-value function considering the current state (\(s_{t}\)) and the action performed (\(a_{t}\)). After learning, the policy $\Pi$ dictating the action to be taken at any given state will be given by: $\Pi (s)=arg\max_{a}{Q(s,a)}$. A full introduction to Q-learning can be found in \cite{sutton1998reinforcement}.

\subsection{Deep Q-Networks (DQNs)}

Deep Q-Networks (DQNs) were introduced in \cite{mnih2015human} and ignited the field of deep Reinforcement Learning for three main reasons: first, DQNs were designed as an end-to-end RL approach when applied to games having only the pixel values and the game score as inputs, thus requiring minimal domain knowledge; second, it used deep convolutional neural networks (CNN) and experience replay to learn a Q-value approximation function successfully; third, it showed very good performance in a wide range of games - in some of them achieving performance higher than human-level - with the use of the same hyper-parameters \cite{li2017deep}. In a nutshell, DQNs are composed of a convolutional neural network to reduce the state space and seek to generalize states, and a Q-learning algorithm for mapping states to actions. Despite their success, DQN training can be computationally inefficient, lacking explainability and hierarchical reasoning.

\subsection{Deep Symbolic Reinforcement Learning (DSRL)} 
DSRL was introduced in \cite{garnelo2016towards}. In DSRL, first, object types are classified with objects having their locations detected by a low-level symbol generator which uses a convolutional auto-encoder and a spectrum comparison technique. Second, a spatio-temporal representation is built of all the objects, using the relative positions of the objects at two consecutive time frames. Third, this spatio-temporal representation is provided to a group of Q-learning functions, one for each pair of type of objects. The update rule for the interaction between objects of types \(i\) and \(j\) is given by:
\begin{equation} \label{eq:DSRL_Q}
Q^{ij}(s_{t}^{ij}, a_{t}) \leftarrow Q^{ij}(s_{t}^{ij}, a_{t}) + \alpha \left[ r_{t+1} + \gamma \max_{a}Q^{ij}(s_{t+1}^{ij},a) - Q^{ij}(s_{t}^{ij},a_{t}) \right]
\end{equation}
where \( \alpha \) is the learning rate, \( \gamma \) is a temporal discount factor, and each state \(s^{ij}_t\)  represents an interaction between object types \(i\) and \(j\) at time \(t\). After learning, the values of the $Q^{ij}$ functions are added, and the action with the largest sum is chosen as shown in Equation \ref{eq:DSRL_a}.
\begin{equation}
a_{t+1} = argmax_{a} \sum_{Q}(Q(s_{t+1},a))
\label{eq:DSRL_a}
\end{equation}

\section{Symbolic Reinforcement Learning (SRL)}
SRL is our implementation of the DSRL model with some simplifications. For ease of analysis, in SRL we separate the Reinforcement Learning problem into three parts: State-Space Representation, Learning, and Decision-Making, as detailed in what follows.

\subsection{State-Space Representation}

\textbf{Naive approach:} The simplest way of representing the state space of an RL environment derived from an image is to associate each possible set of pixel values with a state. A black and white image is composed of a matrix of pixels that can assume values from 0 to 255. In a black and white image where \(n\) is the number of rows and \(m\) is the number of columns of pixels in the image, the number of possible states will then be \(256 ^{n*m}\). 
Such a very large state space presents obvious difficulties for learning. In this setting, each combination of assignments of values to the pixels in the image can be considered a relevant state for the purpose of RL.


\textbf{CNN approach:} In order to reduce the number of states in an image, the use of a convolutional neural network (CNN) was proposed in \cite{DBLP:journals/corr/MnihKSGAWR13}, leading to Deep Q-networks. DQNs seek to generalize the states provided by pixel values using learned filters and a fully-connected feedforward neural network as a regressor for the Q-function value approximation. However, the states generalized by the CNN are by construction dependent on all the pixels in the raw image, which can hinder generalization across different images. In \cite{su2017one}, it is shown for example, that varying one pixel value can cause the classification of a CNN to change, or in the case of DQN, the state to change. Furthermore, the states are not fully translation-, rotation-, and size-invariant, requiring additional manipulation to become more robust, as shown in \cite{jaderberg2015spatial}. States generalized by CNNs do not carry explicit information about the objects' relative positions, and are in general restricted by the fixed dimensions of the input image. Experiments reported in \cite{garnelo2016towards} and reproduced here will highlight this problem in the context of Q-learning.

\begin{figure}[h]
\centering    
\includegraphics[width=0.7\textwidth]
{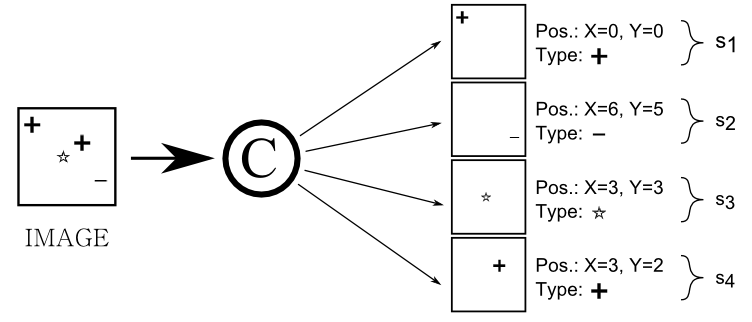}
\caption{The letter C represents the tasks of object recognition and object type assignment typically carried out with the use of a convolutional neural network \cite{garnelo2016towards}. In our approach, such object types and their location are seen as sub-states of the state of an image.}
\label{fig:SR3}
\end{figure}

\textbf{Symbolic approach:} Another way to compress and generalize the number of states in an image that does not incur the above loss of information is to recognize such objects (possibly using a CNN) and represent them by symbols with their respective locations. In the DSRL approach of \cite{garnelo2016towards} this is done by using a low-level symbol generation technique, denoted by the letter C in Figure \ref{fig:SR3}, which illustrates the process. An advantage of DSRL, therefore, is that now the state-space can be represented not by the combination of all pixel values or learned filters but by object types (e.g. `+' and `-' as in Figure \ref{fig:SR3}) and their positions (X and Y numerical values in Figure \ref{fig:SR3}), represented in symbolic form.

\begin{figure}[h]
\centering    
\includegraphics[width=0.6\textwidth]
{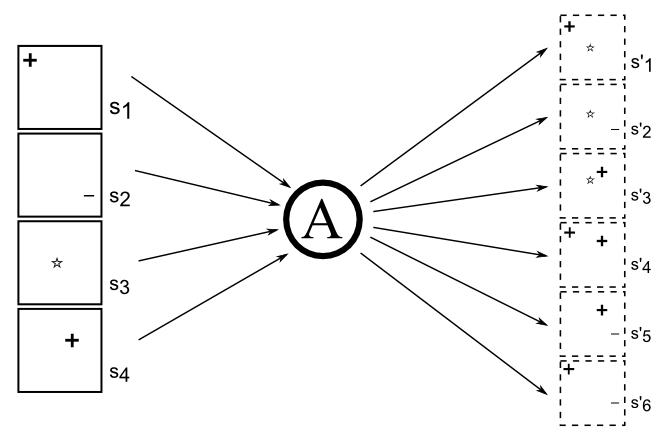}
\caption{Each sub-state can be combined with certain other sub-states to form a richer, and hopefully more meaningful symbolic representation for the learning task at hand. We call this process an abstraction, here denoted by the letter \textbf{A}. In this particular example, \textbf{A} simply calculates the relative position of each object w.r.t. every other object. The boxes drawn using dotted lines indicate that the positions are relative, while the boxes drawn using solid lines indicate that the positions are absolute.}
\label{fig:Abstraction}
\end{figure}

The use of a symbolic approach allows one to separate the detection and classification of objects in an image, which have been shown to work well with the use of CNNs and other deep learning and hybrid approaches \cite{chen2016infogan, garnelo2016towards, li2016relief, salvador2016faster, ijcai2017-221}, from the ways that an agent may learn and reason about its environment. In our current implementation of SRL, used to obtain the results reported in this paper, the object detection and classification part is not included. Instead, we focus on the agent's learning and reasoning. Each object's type and position is provided directly to the system in symbolic form as shown on the right-hand side of Figure \ref{fig:Abstraction}, where we use the letter A to denote \emph{abstraction}. Each box in Figure \ref{fig:Abstraction} can be seen as a sub-state, which can in turn be composed with other sub-states in different ways depending on the choice of abstraction. 

Sub-states that are represented only by a single object often do not carry enough relevant information when it comes to reasoning about the properties of the problem at hand such as size invariance. In the experiments carried out in this paper, which use the same game as used in \cite{garnelo2016towards} where a single agent represented by a star is expected to collect all the positive objects in the environment while avoiding any negative objects, the four objects shown in Figure \ref{fig:SR3} are combined to create the six abstract sub-states shown on the right hand side of Figure \ref{fig:Abstraction}. 

We have observed empirically that the choice of abstraction can facilitate learning considerably by representing the relevant parts of the environment in which an agent is expected to focus. Although it is likely that the \emph{best} abstraction will depend on the problem at hand, and therefore on background knowledge about the problem domain, in what follows we identify some generally desirable (common-sense) abstractions, such as the relative distance between objects, which allows learning to scale well to larger environments, and which can be implemented still by a model-free variation of Q-learning. 

Both in DSRL and SRL, abstraction is carried out by the combination two by two of sub-states. In DSRL, this is done as shown in Figure \ref{fig:Abstraction} by taking the relative position of each object w.r.t. every other object. In SRL, the same is done but then, in addition, any state where an agent (the star-shaped object) is not included is removed (states s$'$4 to s$'$6 in the example of Figure \ref{fig:Abstraction}). This is simply because any state without an agent is obviously irrelevant to our game. In general, ideally, the system would know how to select only sub-states which are relevant to solving the problem. 

An alternative choice of abstraction would essentially dictate how general or specific the resulting state-space should be in relation to the learning task at hand. We leave such an analysis of generality vs. specificity as future work. Notice that the outcome of abstraction in SRL or DSRL is different from the outcome of a neural network-based approach such as DQN. While a CNN is capable of distinguishing one state from another, (symbolic) abstraction is capable of creating entirely new states from the original input. It is likely however that such abstraction can be achieved by stacking end-to-end differentiable neural networks, although this too is left as future work.

In summary, several advantages can be found in the symbolic way of representing the world:

(a) It simplifies the problem by reducing the state-space and limiting the number of relevant objects in each sub-state;\\
(b) It is compositional, allowing the state-space to be represented in different ways which may facilitate learning;\\
(c) Sub-states can reappear at different regions of an image thus contributing to better generalization;\\
(d) It is independent of the size of the images; once a task is learned it should scale to larger environments;\\
(e) It should be possible at least in principle to learn relations among objects.\\

\subsection{Learning}

\begin{figure}[h]
  \centering
  \includegraphics[width=1\textwidth]
  {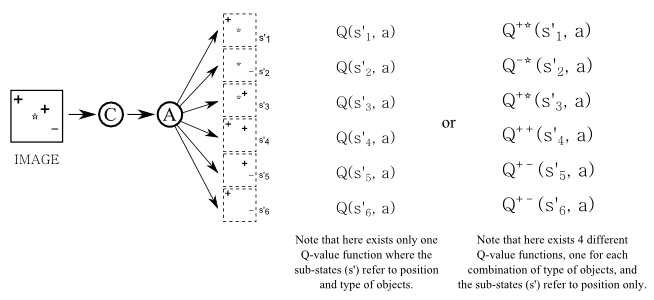}
  \caption{Two options for encoding abstract sub-states into a Q-value function: use a single Q-value function or use as many as the number of pairs of object types. By choosing the latter, another choice will be required on how to use the Q-value functions for decision making.}
  \label{fig:POMDPs}
\end{figure}

\textbf{Learning with multiple sub-states:} A potential drawback of having sub-states is that, instead of solving one Markov Decision Process (MDP), our learning algorithm now has to solve \(N\) Partially Observable Markov Decision Processes (POMDPs), where \(N\) is the number of abstract sub-states (s'). In DSRL and SRL, Q-Learning was chosen for this task, in which case there are two options: use a single Q-value function which embeds the type and position of objects in its sub-states (s'), or use one Q-value function for each pair of object types, embedding only their position in the sub-states. The alternatives are illustrated in Figure \ref{fig:POMDPs}. In DSRL and SRL, the second option was chosen, since having separate Q-value functions should make learning from each specific combination of object types easier. 

In summary, when abstractions (s') are used in the place of entire images, an agent is required to learn in a partially observable environment. In such cases, no guarantees exist of finding an optimal policy, as in the case of a fully-observable environment. Nevertheless, it should still be posible for the learning algorithm to find optimal policies for sub-tasks which may be described fully within the sub-spaces. These sub-tasks often have shorter horizons and may appear more than once, making learning in such cases converge faster, and hopefully helping an agent to succeed in its main task.\footnote{An example would be playing the Atari game of Breakout by focusing first  only on how the paddle moves left or right as a result of the corresponding key strokes, then focusing on certain objects only, such as the paddle and the ball, and how to hit the ball every time, and so on. At some point, e.g. to end the game with a win, an agent could choose to combine all sub-states in its abstraction to form the full image and solve the problem as an MDP. That would require changing the focus from paddle and ball only to the entire state space and probably considering longer horizons, although at this point fewer objects will have been left, and therefore the state space will be smaller than at the start of the game.}

\textbf{Rewarding multiple sub-states:} In Q-learning, a reward is obtained when an action takes place in a given state. An alternative is to associate rewards with the resulting state that an action may produce, or even with reaching a state independently of the action. In SRL with the use of multiple Q-learning functions, therefore, most of the time that a reward is obtained it will be impossible to know to which sub-states this reward corresponds. An agent will perform poorly if rewards are associated with sub-states that are not relevant, or are not the \emph{cause} of the reward.\footnote{For example, in the game of Chess it may be counter-productive to reward for wining the game the positions of the pieces that do not take any part in the King's checkmate.} 

In SRL, the same rewards go into the updates of all the Q-value functions. As discussed, this is undesirable because certain sub-states may be rewarded for an unrelated action. We have confirmed this empirically by creating an example where rewards are assigned incorrectly, leading to low performance at test time. SRL+CS will address this problem by limiting reward updates to specific sub-states based on valid object interactions, as detailed in the next section.



\subsection{Decision Making}

Another issue that arises when considering multiple sub-spaces is how to aggregate the Q-value functions learned in order to make a decision, i.e. choose a single action at a given state.

DSRL selects the highest value among the sum of each Q-value function, as shown in Equation \ref{eq:DSRL_a}. In SRL, we use the same approach. However, this may not work in all situations because the actions associated with the Q-value functions may be contradictory, especially as the number of Q-value functions (i.e. objects) increases. Simply adding the Q-values seems insufficient to guarantee that \emph{argmax} will maximize the accumulated future reward. 

Another way to aggregate the Q-value functions would be to simply take the \emph{argmax} among all the Q-value function values. This does not work well either because it ignores Q-value functions with lower values, which may be relevant nevertheless (e.g. for learning from multiple sub-states). SRL+CS will address this problem by assigning a degree of importance to the different Q-functions, as detailed in the next section.

\section{Symbolic Reinforcement Learning with Common Sense (SRL+CS)}

Symbolic Reinforcement Learning with Common Sense is based on SRL with two modifications, one in the learning and another in the decision-making part, with the objective of solving the two issues identified in the previous section: assigning rewards and aggregating Q-values adequately. In what follows, first, we formalize the state-space representation, then we present the two modifications made in the learning and decision-making algorithms.

\textbf{STATE-SPACE REPRESENTATION:}

SRL+CS creates a state-space exactly as done by SRL. Formally, it builds a representation of the world (abstraction) by creating sub-states \(s^{k}\) which consist of the relative position of an agent w.r.t. an object. Each sub-state represented by this pair (agent, object) is denoted by \(k\). Let agent \(m\) and object \(n\) have absolute positions \((x^m,y^m)\) and \((x^n,y^n)\), respectively. Then, \(s^{k}\) is a tuple \((x^m - x^n, y^m - y^n)\).  

\textbf{LEARNING:}

In order to assign rewards adequately, SRL+CS restricts the Q-value function updates to specific sub-states, based on the interactions between an agent and the objects. Specifically, when a reward that is non-zero is received, only the Q-value function with a sub-state \(s^{k}\) with a value of (0,0) is updated. Recall that a value of (0,0) indicates that an agent is in direct contact with the object. Equation \ref{eq:SRL+CS_learn} shows the Q-value function update:

\begin{equation} 
Q^{ij}(s^{k}_{t}, a_{t}) \leftarrow Q^{ij}(s^{k}_{t}, a_{t}) + \alpha \left[ r_{t+1} + \gamma \max_{A}Q^{ij}(s^{k}_{t+1},A) - Q^{ij}(s^{k}_{t},a_{t}) \right]
\label{eq:SRL+CS_learn}
\end{equation}

where \( \alpha \) is the learning rate, \( \gamma \) is the temporal discount factor, and \( A \) is the set of all possible actions for that state \( s^{k}_{t+1} \). Notice that more than one update to the Q-value function with indices \(i\) and \(j\) can take place (i.e. for different sub-states) at each time point \(t\) due to the presence of multiple objects of the same type in the image. Equation \ref{eq:SRL+CS_learn} is very similar to the standard Q-learning approach, with the difference that now usually the update does not occur once, but for all sub-states (k). When the reward is different from zero, an update occurs only for the sub-state with value (0,0). Intuitively, this is expected to indicate \emph{where a reward has been coming from}, given a set-up consisting of multiple sub-states.

\textbf{DECISION MAKING:}

In order to aggregate Q-values adequately, SRL+CS assigns a degree of importance to each Q-value function, based on how far the objects are from an agent. The calculation of the distances from an agent to the objects is straightforward given the above choice of symbolic state-space representation. This allows one to give a priority to the decisions that are spatially close and therefore generally more relevant. It should also help an agent concentrate on a relevant subtask; in the case of the game used in the experiments that will follow, a relevant sub-task may be to collect the positive objects which are above or below. This idea is in line with that of a common sense \emph{core} whereby our actions are influenced mostly by our surroundings \cite{lake2017building}, giving more importance in general to the objects that are nearer or can be seen. 

Formally, the next action is chosen by Equation (\ref{eq:SRL+CS_act}), where \( d^k_{t} \) is the Euclidean distance between the agent and the object's position at time \(t\).

\begin{equation} \label{eq:SRL+CS_act}
a_{t+1} = arg\max_{A} \left[ \sum^{k} \dfrac{Q^{ij}(s^{k}_{t},A)}{ (d^{k}_{t})^2} \right]
\end{equation}

Notice the similarity with Equation \ref{eq:DSRL_a}. As in SRL, Q-values are summed and the action associated with the function with the largest value is taken. It is worth pointing out that, as before, this is not guaranteed to work in all situations, although it works very well in most situations of the game used in our experiments, reported in the next section.
 

\section{Experimental Results}
We have implemented a game, also used in \cite{garnelo2016towards}, to evaluate and compare the performances of the following RL algorithms: Q-Learning, DQN, DSRL, SRL and SRL+CS, as introduced and discussed in the previous sections. In this game, an agent (a star-shaped object) can move up, down, left or right to collect as many positive objects as possible, denoted by a plus sign, while avoiding as many negative objects as possible, denoted by a minus sign. The objects are at fixed positions. Every time that the position of the star-shaped agent coincides with that of an object the object disappears, yielding a reward of 1 in the case of positive objects and -1 in the case of negative objects. A set of bricks restricts the passage of the agent, as shown in Figure \ref{fig:Env567}. The game finishes when all positive objects are collected or after 100 movements of the agent. 

The environment is fully-observable (the agent can see all the objects), sequential, static, discrete and finite, with no model of the world provided in advance.

All algorithms (Q-Learning, DQN, DSRL, SRL and SRL+CS) used an \(\epsilon\)-greedy exploration strategy with a 10\% chance of choosing a different action at random during training. In addition, actions were chosen randomly when two or more actions shared the same Q-value, i.e. no predefined preference was assigned to any action. The learning rate (alpha) was set to 1.0 since the state transitions were deterministic, and the discount-factor rate was set to 0.9 as a standard value.

The metric chosen for measuring the performance of the star-shaped agent was a score calculated as the sum of accumulated rewards. This tells us how much an agent has learned to move correctly by collecting positive objects while avoiding negative ones.

\subsection{Training in Larger State-spaces}

\textbf{Experiment 1 - Increasing the Size of the Environment.}

\begin{figure}[h]
  \centering
  \null\hfill
    \subfloat[Grid 3x3]{
    \includegraphics[width=0.15\textwidth]			     {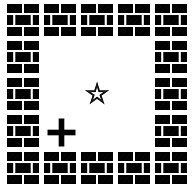}
  \label{fig:Env5}}
  \hfill
    \subfloat[Grid 5x5]{
    \includegraphics[width=0.2\textwidth]
    {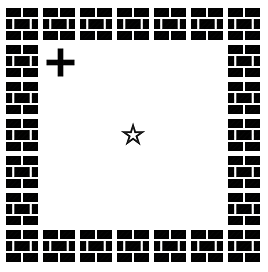}
  	\label{fig:Env6}}
  \hfill
    \subfloat[Grid 7x7]{
    \includegraphics[width=0.25\textwidth]
    {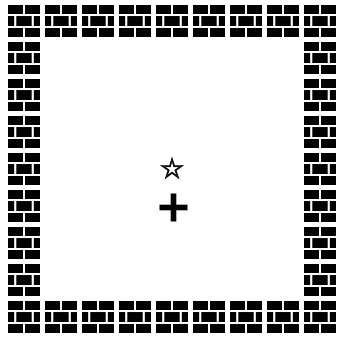}
  \label{fig:Env7}}
  \null\hfill
  \caption{Environments of increasing sizes used in Experiment 1 with positive objects only.}
  \label{fig:Env567}
\end{figure}

In this first experiment, agents were trained on the three grids shown in Figure \ref{fig:Env567} of increasing sizes. Starting from the center of the grid, an agent has to collect a positive object placed at a random location within the grid. No negative objects were used in this experiment. 

In SRL (and DSRL and SRL+CS), an abstraction is responsible for building a state space that takes into consideration the relative positions of the objects. As a result, an increase in the size of the environment should not imply the same increase in the size of the state-space. If the relative position between agent and object is the same in both the small and large grids, the agent's performance should be roughly the same regardless of the size of the environment. This is not normally the case with the use of Q-learning.

\begin{figure}[h]
  \centering
    \subfloat[Results of grid 3x3]{
    \includegraphics[width=0.32\textwidth]
    {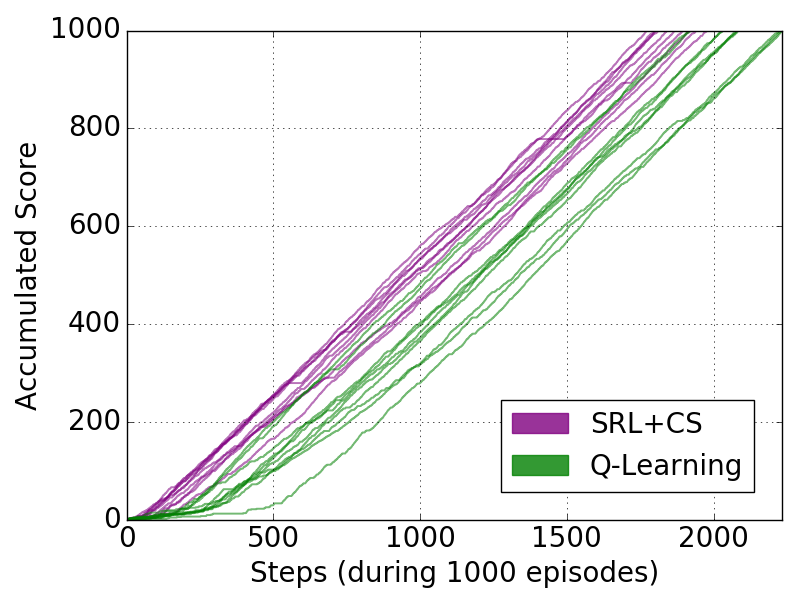}
  \label{fig:env5_r}}
    \subfloat[Results of grid 5x5]{
    \includegraphics[width=0.32\textwidth]
    {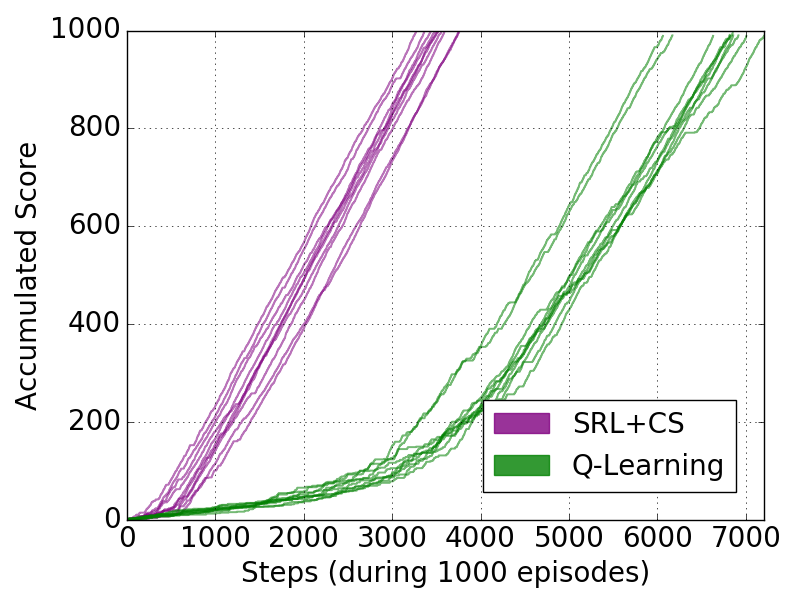}
  \label{fig:env6_r}}
    \subfloat[Results of grid 7x7]{
    \includegraphics[width=0.32\textwidth]
    {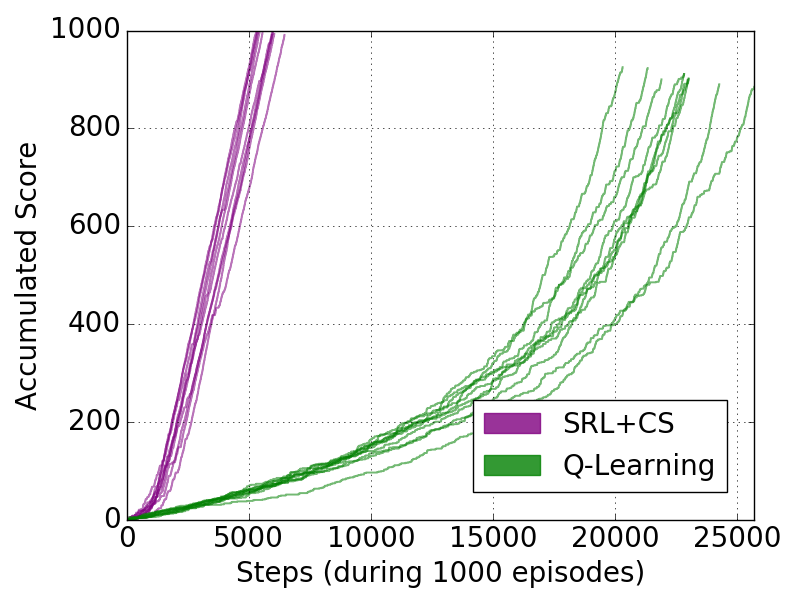}
  \label{fig:env7_r}}
  \caption{Experiment 1 - These three plots, each containing 10 runs of 1000 games (each game is also referred to as an \emph{episode}; each \emph{step} is one movement of the agent in the grid) show that the symbolic approach (in this case SRL+CS) learns to collect the positive objects much faster than Q-learning as the size of the grid increases; SRL obtained very similar results to SRL+CS, but these were not plotted to avoid occlusion.}
  \label{fig:Results_Env567}
\end{figure}

Figure \ref{fig:Results_Env567} shows, in each plot, the accumulated scores obtained over 10 runs of 1000 games each, using both SRL+CS and Q-learning. It can be seen that, as the size of the grid increases, Q-learning requires many more training steps (i.e. movements of the agent within the grid) than SRL+CS. The results confirm those reported in \cite{garnelo2016towards}.

\subsection{Training and Testing in Deterministic and Random Configurations}

\begin{figure}[h]
\centering
\subfloat[Deterministic Configuration]{
\includegraphics[width=0.32\textwidth]{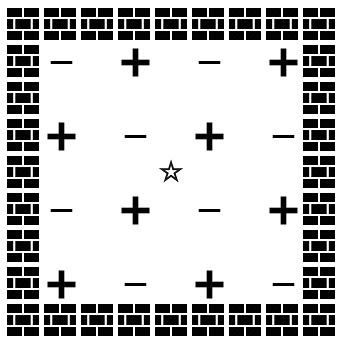}
\label{fig:Env10}}
\qquad
\subfloat[Random Configuration]{
\includegraphics[width=0.32\textwidth]{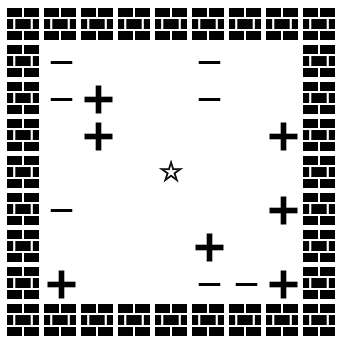}
\label{fig:Env11}}
\qquad
\caption{Two examples of initial configurations of the environment used in Experiments 2, 3 and 4, where each episode starts with eight (a) and seven (b) objects of each type.}
\label{fig:Env1011}
\end{figure}

We now run three experiments using the two configurations shown in Figure \ref{fig:Env1011} where the environment is initialized at each episode either deterministically or at random with the same number of both positive and negative objects. These configurations are the same as used in the experiments reported in \cite{garnelo2016towards}.

In addition to the \emph{accumulated scores} metric already used to measure agent performance, here we also use the percentage of positive objects collected over the total number of objects collected: a value of 100\% shows that an agent has collected only positive objects, but not necessarily all positive objects.

In \cite{garnelo2016towards}, for some experiments, only this second metric is used. We argue that it should be used in connection with the accummulated scores, as done here, as otherwise it may boast the performance of a very conservative agent, one which collects very few positive objects. 

\textbf{Experiment 2 - Training in a Deterministic Configuration.}\\
In this experiment, starting from the center of the grid, an agent has to collect positive objects while avoiding negative ones, all positioned in the grid by following a deterministic pattern as illustrated in Figure \ref{fig:Env10}.

\begin{figure}[h]
\centering
\subfloat[Accumulated Scores for 10 runs of 1000 episodes each for each algorithm.]{
\includegraphics[width=0.45\textwidth]{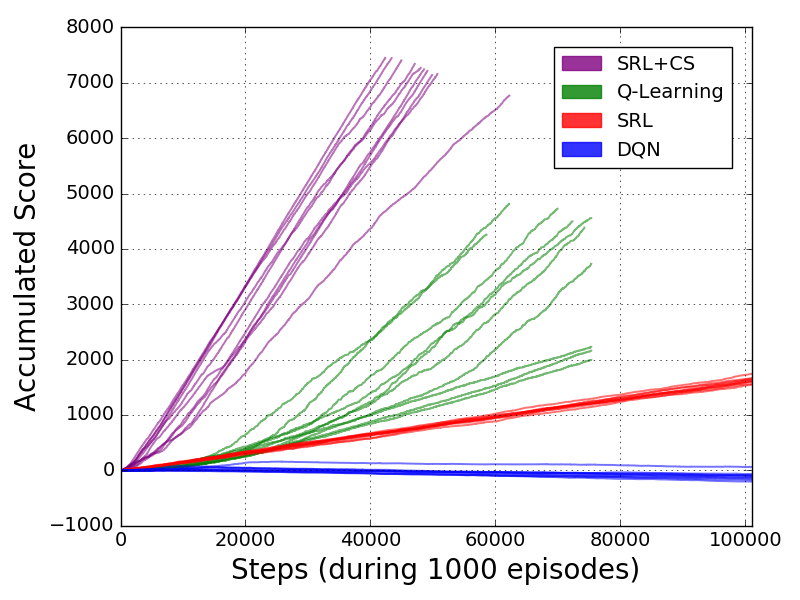}
\label{fig:subfig5}}
\qquad
\subfloat[Rolling mean (using a window of ten episodes) of the average percentage of positive objects collected over 10 runs. ]{

\includegraphics[width=0.45\textwidth]{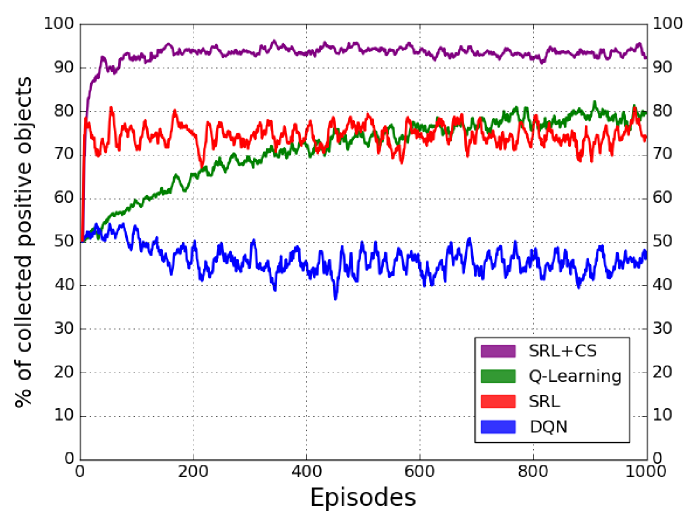}
\label{fig:subfig6}}
\qquad
\caption{Experiment 2 - Fig. 7(a) shows that SRL+CS learns faster to collect positive objects and avoid negative objects. Some Q-learning curves start to show the same slope as SRL+CS but much later. Fig. 7(b) shows that the rate of positive objects collected using SRL+CS is more than 90\%. Notice that SRL+CS cannot reach 100\% as an average due to the effect of 10\% random exploration built into the learning algorithm.
}
\label{fig:Result_Env10}
\end{figure}

The results of this experiment, shown in Figure \ref{fig:Result_Env10}, indicate that SRL+CS learns a policy which is better than that chosen by any of the other algorithms, faster than any of the other algorithms. SRL+CS collects an average rate of more than 90\% of positive objects. DQN did not achieve a reasonably good performance in our experiments, probably because it requires a large number of hyper-parameter fine tunning, which was not done extensively. In this same task, results reported in \cite{garnelo2016towards} show that DQN can achieve a rate of almost 100\% of positive objects collected, and that DSRL can achieve a rate of 70\% of positive objects collected, although accumulated scores were not reported. 

\textbf{Experiment 3 - Training in a Random Configuration.}\\
In this experiment, the positions of the objects to be placed on the grid are chosen randomly at the start of each episode, except for the agent's position, which is always the center of the grid, as exemplified in Figure \ref{fig:Env11}.

\begin{figure}[h]
\centering
\subfloat[Accumulated Scores for 10 runs of 1000 episodes each for each algorithm. 
]{
\includegraphics[width=0.45\textwidth]{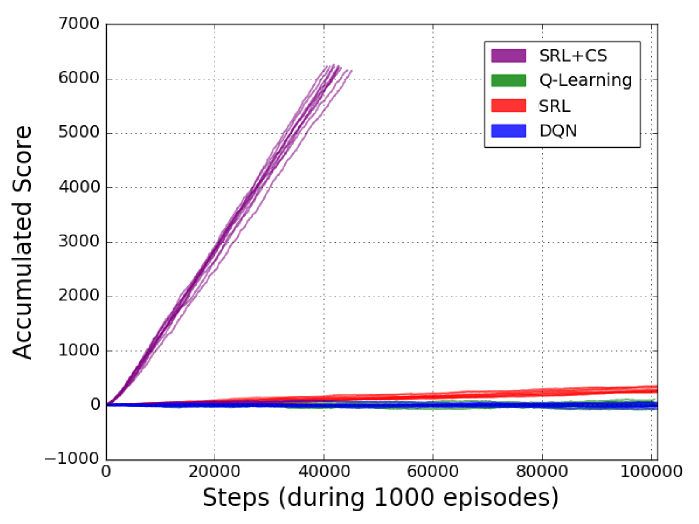}
\label{fig:subfig7}}
\qquad
\subfloat[Rolling mean (using a window of ten episodes) of the average percentage of positive objects collected over 10 runs. 
]{
\includegraphics[width=0.45\textwidth]{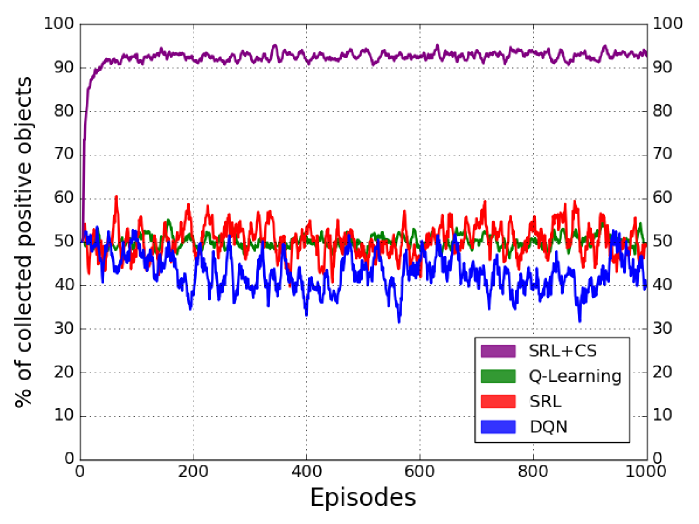}
\label{fig:subfig8}}
\qquad
\caption{Experiment 3 - Figs. 8(a) and 8(b) show that only SRL+CS learns an effective policy for the random environment. This indicates that, independently of the configuration of the environment, SRL+CS learns how to interact with the objects and to move correctly in order to collect as many positive objects as possible avoiding negative objects.}
\label{fig:Result_Env11}
\end{figure}

The results presented in Figure \ref{fig:Result_Env11} show that only SRL+CS is capable of learning an effective policy in this case. It suggests that SRL+CS can learn certain sub-tasks, despite the randomness of the environment, unlike the other algorithms. SRL+CS must have learned sub-tasks independently of where the objects were located, for example,
\emph{move up if a positive object is above}, even though SRL+CS is not provided with any explicit knowledge.

\textbf{Experiment 4 - Training in a Deterministic Configuration and Testing in a Random Configuration (Zero-shot Transfer Learning).}\\

In this experiment, we evaluate the system's ability to re-use knowledge learned in a situation (Fig. \ref{fig:Env10}) in a different yet related situation (Fig. \ref{fig:Env11}). All systems were trained on the deterministic configuration and tested on the random configuration. During testing, there was no further learning involved; hence, there was no exploration either. 

\begin{figure}[h]
\centering
\subfloat[Accumulated Scores for 10 runs of 1000 episodes each, for each algorithm. (*) Results for DQN and DSRL are not available.
]{
\includegraphics[width=0.45\textwidth]{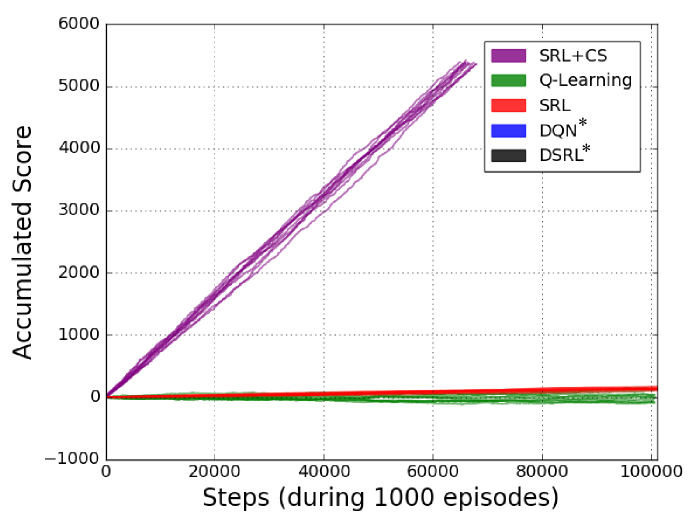}
\label{fig:score_test_11}}
\qquad
\subfloat[Rolling mean (using a window of ten episodes) 
of the average percentage of positive objects collected over 10 runs. The results reported for DSRL and DQN are obtained from \cite{garnelo2016towards}. 
]{
\includegraphics[width=0.45\textwidth]{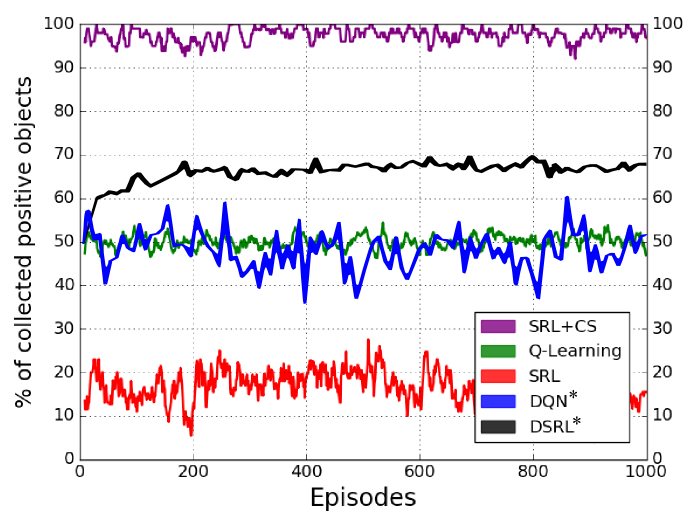}
\label{fig:percent_test_11}}
\qquad
\caption{Experiment 4 - In this zero-shot transfer learning experiment, SRL+CS does not receive any training in the random environment task, but is the only algorithm to solve the task.}
\label{fig:Result_Env11_Test}
\end{figure}

The results of this experiment (Figure \ref{fig:Result_Env11_Test}) show that only SRL+CS is able to transfer its learning correctly to a similar situation; SRL+CS collects almost 100\% of the positive objects in the random environment. Such results are impressive and deserve further investigation. Figure \ref{fig:percent_test_11} shows that SRL collects only 15\% of positive objects, but SRL achieves a higher score than Q-Learning (c.f. Figure \ref{fig:score_test_11}). This is because in many of the runs SRL did not collect any object at all, making the average of the percentage of positive objects collected lower than a random walk (where the same amount of negative and positive objects are collected). DQN and Q-learning were unable to transfer knowledge learned in the deterministic environment to the new situation. Additionally, DSRL was able to collect only 70\% of positive objects.

We also ran other experiments training and testing SRL+CS in a number of different game configurations, achieving similar results as reported here (apart from one case where the possible movements of the agent are restricted by the environment, e.g. in a 3x2 environment with one positive and one negative object). This is discussed next.

\subsection{Further Transfer Learning Experiments}

\begin{figure}[h]
\centering
\subfloat[Train]{
\includegraphics[width=0.2\textwidth]			     {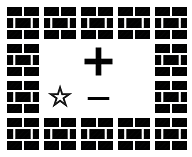}
\label{fig:Env1}}
\hfill
\subfloat[Test]{
\includegraphics[width=0.2\textwidth]
{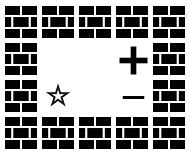}
\label{fig:Env2}}
\hfill
\subfloat[Test]{
\includegraphics[width=0.2\textwidth]
{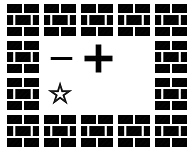}
\label{fig:Env3}}
\caption{Initial configurations used in Experiments 5 and 6.} 
\label{fig:Env123}
\end{figure}

Finally, we run two transfer learning experiments using the three configurations shown in Figure \ref{fig:Env123}. For the purpose of highlighting in the plots if a negative object is collected, in these experiments a different reward function is chosen. If a negative object is collected, a reward of -10 is obtained, and if a positive object is collected, a reward of +1 is obtained. We use 10 steps, instead of 100, as the maximum number of steps in each episode, since the environments are much smaller now.

\begin{figure}[h]
\centering
\includegraphics[width=0.8\textwidth]
{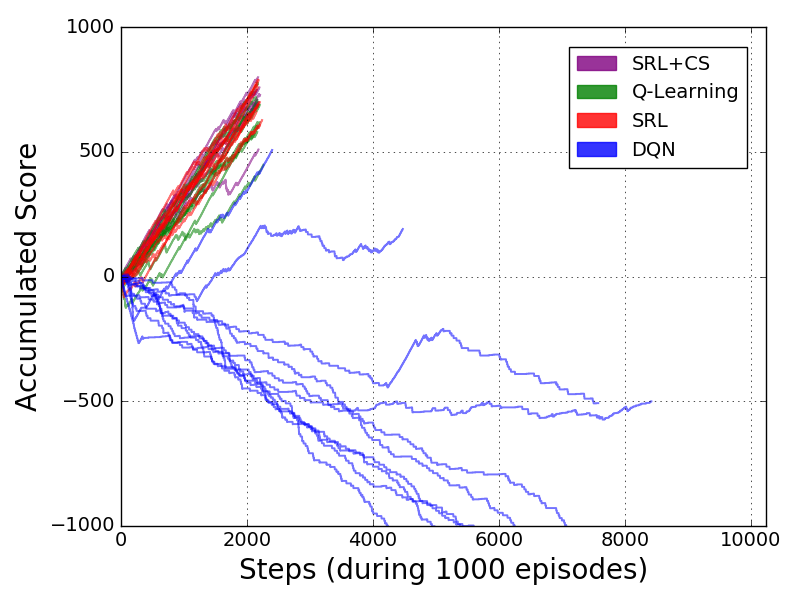}
\label{fig:env5_r}
\caption{Scores for training in configuration (a), showing 10 runs for each algorithm, each run containing 1000 episodes. For all algorithms, apart from DQN, agents have learned to collect the positive object and to avoid the negative object. DQN, however, is quite inconsistent, although the best DQN run, whose model is selected for the rest of the experiments, achieves performance similar to the other algorithms. 
} 
\label{fig:Result_Env1}
\end{figure}

Figure \ref{fig:Env123} shows the three configurations used. The systems were trained in configuration (a) and tested in configurations (b) and (c). Figure \ref{fig:Result_Env1} shows the algorithms' training performances in configuration (a). Out of the 10 runs for each algorithm, the best models (the ones achieving the highest scores) were selected for testing in configurations (b) and (c). Testing in these configurations was done once per algorithm since the configurations were deterministic and there was no further learning.

\begin{figure}[h]
\centering
\includegraphics[width=0.65\textwidth]{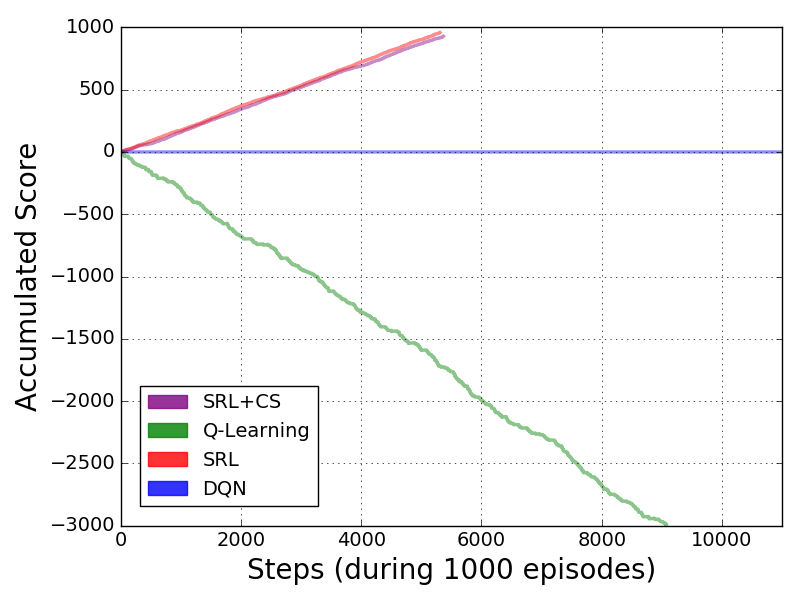}
\label{fig:env6_r}
\caption{Experiment 5 - The symbolic algorithms are able to transfer learning correctly from environment (a) to environment (b), while Q-learning behaves randomly, and DQN never collects any object.}
\label{fig:Result_Env12}
\end{figure}

\textbf{Experiment 5 - Transferring Undefeasible Knowledge.}\\
In this experiment, after training in configuration (a) the agents were tested in configuration (b). Figure \ref{fig:Result_Env12} shows that DQN is unable to collect the positive objects, Q-learning shows a random behavior, and the symbolic algorithms were able to generalize to the new situation, thus avoiding the negative object while collecting the positive object. By inspecting configurations (a) and (b),  a reasonable assumption is that the symbolic approaches can learn \emph{to move up and to the right}, although no explicit knowledge is used (recall that Equations \ref{eq:SRL+CS_learn} and \ref{eq:SRL+CS_act} are the only changes made to Q-learning). Comparing configurations (a) and (c) though, it is clear that knowledge learned in (a) needs to be revised in (c), as discussed next. 

\begin{figure}[h]
\centering
\includegraphics[width=0.65\textwidth]
{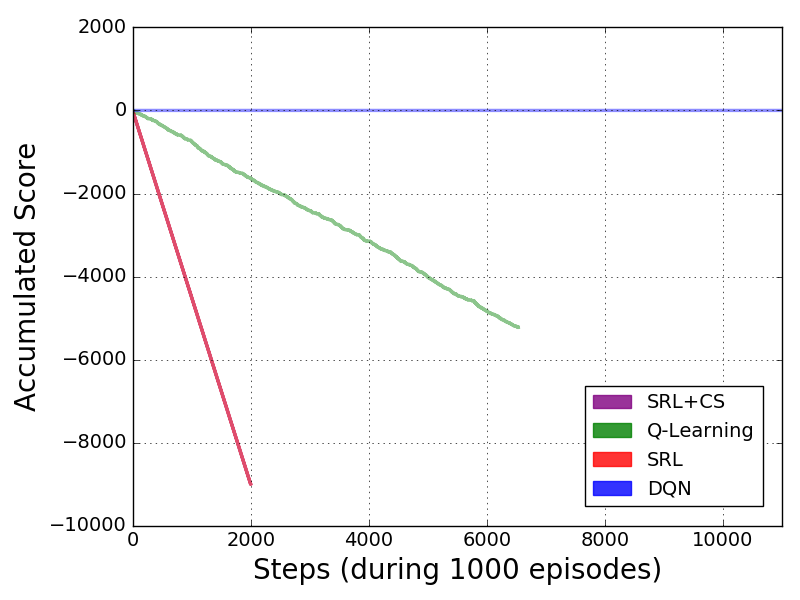}
\label{fig:env7_r}
\caption{Experiment 6 - It this case of negative transfer, the symbolic algorithms fail, while Q-Learning behaves randomly, performing better than SRL and SRL+CS as a result, and DQN never collects any object. In this experiment, SRL and SRL+CS share exactly the same behaviour of collecting the negative object before collecting the positive object.}
\label{fig:Result_Env13}
\end{figure}

\textbf{Experiment 6 - Negative Transfer.}\\
In this experiment, following training in configuration (a), the systems are tested in configuration (c). Figure \ref{fig:Result_Env13} shows the results. Now, DQN does not collect any object, as before, Q-Learning walks randomly, as before, but SRL and SRL+CS collect the negative object before collecting the positive object in every episode.

Experiment 6 was the only case where SRL+CS failed to generalize. By inspecting configurations (a) and (c), it is clear that if an agent learns \emph{to move up and to the right} then it shall collect the negative object always in configuration (c). This is a case of negative transfer, where the system has no means of revising the knowledge learned. Notice that the problem is not in the decision-making, which should distinguish positive and negative objects, but in the learning phase. Symbolic RL algorithms use Q-learning, which learns from (state, action) pairs. This may not be desirable, as seen here, leading to poor generalization. In other words, Q-learning cannot generalize over states only. Because of this, an agent will not learn to simply \emph{avoid a negative object} (recall that during learning this agent did not encounter a situation where it was below a negative object and executed the action of moving up). In many real-world situations, it should not matter how a state is reached, and the simple fact of being in a state can be sufficient to make a decision on a relevant action. This limitation of the generalization capacity of Q-learning is well-known.

\section{Conclusion and Future Work}
In this paper, we have presented a symbolic RL algorithm with features of common sense (SRL+CS) which can outperform Q-learning, DSRL and DQN in environments with deterministic and random configurations, and at transfer learning.
Building on the knowledge representation proposed by deep symbolic RL, SRL+CS can make two simple modifications to the standard Q-learning algorithm, implementing common sense features such that attention can be focused normally on the objects that are of most relevance to an agent. 

SRL+CS is model free: the choice of an appropriate abstraction for the representation of the state space allows for simple changes to learning and decision making which enhance the generalization capacity of RL. The experiments reported in this paper were also run with only one of the two modifications made at a time. Although each modification has led to an improvement in performance, only the combination of the two has produced the high levels of performance seen in all the tasks. 

The results reported reiterate the value of the symbolic approach to RL, which through the use of sub-state abstractions can reduce the size of the state space considerably. However, we have shown that two important issues arise: the assignment of rewards to sub-states during learning and the aggregation of Q-values for decision making. SRL+CS addresses both issues adequately. 

Various features of common sense have been argued to be relevant for achieving an adequate balance between generalization and specialization \cite{davis2015commonsense, mccarthy1960programs}. Symbolic RL offers the possibility of adding to standard RL features of common sense which go beyond those considered in this paper. Despite the impressive results of SRL+CS at zero-shot learning, the experiments reported also highlight the limitations of Q-learning at generalizing from state-action representations. As future work, therefore, we aim to build an end-to-end SRL+CS architecture capable of learning common sense domain knowledge and applying it to solving more complex problems. A promising direction for further investigation is the interplay between SRL+CS and related work on the integration of planning and RL \cite{Kansky2017, sutton1990first, vanseijen2015deeper} and Relational RL \cite{dvzeroski2001relational, nickles2011integrating}.


\end{document}